\documentclass{article}

\PassOptionsToPackage{numbers, compress}{natbib}

     \usepackage[preprint]{neurips_2020}

\usepackage[utf8]{inputenc} 
\usepackage[T1]{fontenc}    
\PassOptionsToPackage{hyphens}{url}\usepackage{hyperref}     
\usepackage{booktabs}       
\usepackage{amsfonts}       
\usepackage{nicefrac}       
\usepackage{microtype}      
\usepackage{graphicx}
\usepackage{amsmath}
\usepackage{caption}
\usepackage{subcaption}
\usepackage{xspace}
\usepackage{listings}
\usepackage{xcolor} 
\usepackage[inline]{enumitem}
\newcommand*{\eg}{e.g.\@\xspace}
\newcommand*{\ie}{i.e.\@\xspace}
\newcommand*{\wrt}{w.r.t.\@\xspace}
\usepackage{wrapfig}
\definecolor{light-gray}{gray}{0.95}
\newcommand{\RNum}[1]{\uppercase\expandafter{\romannumeral #1\relax}}
\usepackage{appendix}
\usepackage{rotating}
\newtheorem{metric}{Metric}[section]
\title{On quantitative aspects of model interpretability}

\author{%
  An-phi Nguyen\\
  IBM Research Z\"urich, ETH Z\"urich\\
  \texttt{uye@zurich.ibm.com} \\
  \And
  Mar\'ia Rodr\'iguez Mart\'inez\\
  IBM Research Z\"urich\\
  \texttt{mrm@zurich.ibm.com} \\
}

\begin{document}

\maketitle

\begin{abstract}
  Despite the growing body of work in interpretable machine learning, it remains unclear how to evaluate different explainability methods without resorting to qualitative assessment and user-studies. While interpretability is an inherently subjective matter, previous works in cognitive science and epistemology have shown that good explanations do possess aspects that can be objectively judged apart from \emph{fidelity}), such as \emph{simplicity} and \emph{broadness}. In this paper we propose a set of metrics to programmatically evaluate interpretability methods along these dimensions. In particular, we argue that the performance of methods along these dimensions can be orthogonally imputed to two conceptual parts, namely the \emph{feature extractor} and the actual \emph{explainability method}. We experimentally validate our metrics on different benchmark tasks and show how they can be used to guide a practitioner in the selection of the most appropriate method for the task at hand. 
\end{abstract}

\section{Introduction}\label{sec:intro}

The issue of interpretability and explainability in machine learning, albeit being a relatively old one~\cite{Thagard1991PhilosophicalExplanation}, has experienced a recent surge in interest from society at large, from researchers to policymakers~\cite{Vishwanath2001TowardMarkets}. An evidence of this increased interest is the amount of review papers that has been written on the topic, \eg~\cite{Guidotti2018AModels,Gilpin2019ExplainingLearning,Carvalho2019MachineMetrics,BarredoArrieta2020ExplainableAI,Tjoa2019AXAI}. Despite the rapid growth of the interpretable machine learning field, it is still unclear how different methods can be \emph{programmatically} compared without the need of qualitative assessments or user-studies. 

A general computational benchmark across all possible methods is unlikely to be possible. After all, interpretability is an inherently \emph{subjective} matter. As \citet{Miller2019ExplanationSciences} stated, explanations are \emph{contextual}. That is, the perceived quality of an explanation is dependent on the (background of the) two interacting agents, \ie the provider and the receiver of an explanation, and the type of information that is of interest to the receiver.

Inspired by previous work from cognitive science and epistemology, in this paper we identify three possible \emph{quantitative} aspects of interpretability and provide a novel set of metrics to measure them. These metrics can guide the practitioner in the selection of the most appropriate method for the task at hand. The metrics  are not meant to replace user-studies, but  to guide the selection of a small subset of explanations to present to participants of an user-study, reducing the overall financial and time burden of such experiments. Using nomenclature introduced by \citet{Doshi-Velez2017TowardsLearning}, our metrics can be classified as \emph{functionally-grounded} metrics, i.e. they rely on some definition of interpretability, rather than on human   experiments.  

A crucial step to define these metrics is the conceptual separation of an interpretability method into a \emph{feature extractor} and the actual \emph{explainability method}. This is consequence of the contextual nature of interpretability: using a natural language analogy, the feature extractor provides the alphabet, while the explainability method provides the grammar.

Our work primarily focuses on  metrics for \emph{model interpretability}, and not data interpretability/analysis. While the two fields may share a great number of tools, the former is interested in understanding how a (generally human-implemented) model will behave on given data. The latter, instead, is interested in understanding the true unknown stochastic process generating the data.

\subsection{Quantitative aspects: simplicity, broadness, and fidelity}\label{sec:quant_asp}

Studies by \citet{Lombrozo2016ExplanatoryInference} show that explanations play an important role in humans' cognitive processes, especially learning and inference. Perhaps not too surprisingly, the explanations generally considered good are \emph{simple} and \emph{broad}, \ie more generally applicable. \citet{Ylikoski2010DissectingPower} provide a more elaborate account of these aspects. In particular, they propose dimensions along which the simplicity and broadness of explanations could be \emph{objectively} assessed: 
\begin{itemize}
    \item \emph{non-sensitivity}: how robust is an explanation to unimportant details;
    \item \emph{factual accuracy}: how detailed is an explanation. An example of a \emph{non-}factually accurate explanation is an idealised one. In a machine learning context, it may be a feature with discretized values;
    \item \emph{degree of integration}: how generally applicable is the explanation;
    \item \emph{cognitive salience}: how easily can an explanation be grasped.
\end{itemize}

These works do not consider \emph{fidelity}, or \emph{faithfulness}, \ie how correct is the explanation. This is simply because they \emph{assume} that the explanations to be evaluated are true, since it would be meaningless, from an epistemologic perspective, to discuss a false explanation. On the other hand, fidelity plays a central role in interpretable machine learning~\cite{Molnar2019InterpretableLearning,Yeh2019OnExplanations,Lipton2016TheInterpretability,AlvarezMelis2018TowardsNetworks}, since often the computed explanations are only an approximation (e.g. surrogate modeling~\cite{Molnar2019InterpretableLearning,Ribeiro2016Classifier,Ribeiro2018Explanations}).


\section{Simplicity, broadness, and fidelity in practice}

While the previously introduced aspects play a crucial role, we believe that \emph{it is not possible} to define an implementation of metrics which can be applied to all the state-of-the-art interpretability methods. This is consequence of the contextual nature of interpretability, which manifests itself in two ways: in the \emph{features} used for the explanation, and in the actual explainability \emph{modality}. For example, in the case of an image classification task, we argue it would be difficult to define metrics applicable to LIME~\cite{Ribeiro2016Classifier} (which assigns attributions to super-pixels), traditional feature attribution methods~\cite{Ancona2018TowardsNetworks} (which assign attribution to individual pixels) or \texttt{MMD-Critic}~\cite{Kim2016ExamplesInterpretability} (which presents explanations in form of examples instead of attributions).

Therefore different metrics need to be defined for the \emph{feature extractor} (Section~\ref{sec:feat_ext}) and the actual \emph{explainability method}. Furthermore, for an explainability method the implementation of the metrics will change according to the explanation modality (Section~\ref{sec:interpret_metrics}).

We emphasize that the feature extractor and the explainability model play \emph{orthogonal roles} in the performance of a method in terms of simplicity, broadness, and fidelity. That is, the quality of an explanation can be easily changed by independently modifying either the features used or the underlying explainability model.

\subsection{Feature Extractor metrics}\label{sec:feat_ext}

Feature extraction is leveraged to create an \emph{interpretable data representation}~\cite{Ribeiro2016Classifier,Sokol2019BLIMEy:LIME}. This is especially true for high-dimensional data which is generally difficult to understand for a human~\cite{Kim2018InterpretabilityTCAV,Marois2005CapacityBrain}. Since data  processing may change the information content of the original samples, we propose to use \emph{mutual information}~\cite{Cover1991ElementsTheory} to monitor simplicity, broadness and fidelity.

\begin{metric}[Mutual Information] Let us denote with $X$ a random variable (r.v.) taking values in the \emph{original} space $\mathcal{X}$, with $Z = g(X) \in \mathcal{Z}$ the corresponding r.v. after feature extraction, and with $Y \in \mathcal{Y}$ the target values (\eg class labels). Simplicity and broadness can be monitored with the \emph{feature} mutual information I(X, Z), while fidelity with the \emph{target} mutual information I(Z, Y).
\end{metric}

Low mutual information can be achieved by using a \emph{non-injective} feature extractor $g(\cdot)$. A non-injective extractor means that some information is discarded (simpler, but less factually accurate explanation) and that multiple original samples are mapped to the same value in the feature space $\mathcal{Z}$ (broader explanations, with high degree of integration). However, a too low feature mutual information would lead to a loss of information about the target $Y$ (because of the data processing inequality~\cite{Cover1991ElementsTheory}). Indeed, we should maintain the target mutual information as high as possible to provide a faithful explanation. A trade-off between the feature mutual information and the target mutual information corresponds to a trade-off between simplicity, broadness and fidelity.

%

\subsection{Metrics for different modalities}\label{sec:interpret_metrics}

\subsubsection{Example-based methods}\label{sec:example_based}

Example-based methods are interpretability methods providing a summarized overview of a model using representative examples, or high-level concepts \cite{AlvarezMelis2018TowardsNetworks,Kim2018InterpretabilityTCAV,Chen2019ThisRecognition,Guidotti2020BlackSpace}.
Given a prediction of interest, an example-based method explains the prediction either directly, providing examples with the same prediction~\cite{Chen2019ThisRecognition}, or counterfactually, by providing examples with a different prediction~\cite{Dhurandhar2018ExplanationsNegatives}.
For example-based methods, we propose the following metrics.

\begin{metric}[Example-based method metrics]\label{metric:example_based} Let us denote with $f: \mathcal{X} \rightarrow \mathcal{Y}$ the model that we want to interpret. To evaluate the explanation for a prediction of interest $y^* \in \mathcal{Y}$ with a set of $N_E$ examples $E = \{\mathbf{x}^E_1,\dots, \mathbf{x}^E_i, \dots, \mathbf{x}^E_{N_E}\}$, the following quantities should be monitored:
\begin{itemize}
    \item \emph{Non-Representativeness}: $
            \frac{\sum_{\mathbf{x} \in E}l(y^*, f(\mathbf{x}))}{N_E} $
    \item \emph{Diversity}: $
            \sum_{\substack{\mathbf{x}_i, \mathbf{x}_j \in E\\\mathbf{x}_i\neq \mathbf{x}_j}}\frac{d(\mathbf{x}_i, \mathbf{x}_j)}{2N_E} $
\end{itemize}
where $d(\cdot,\cdot)$ is a distance function defined in the input space $\mathcal{X}$.
\end{metric}

The non-representativeness is mainly a measure of the fidelity of the explanation. High non-representativeness, however, can also be indicative of factual inaccuracy. A highly diverse set of examples demonstrates the degree of integration of the explanation. The cognitive saliency, and ultimately the simplicity, of the explanation is simply encoded in the number $N_E$ of examples used: the least number of examples, the easier it is for a human to process the explanation~\cite{Marois2005CapacityBrain}.

\subsubsection{Feature Attributions}\label{sec:feat_attr}

Feature attribution methods seek to assign an \emph{importance value} to each feature depending on its contribution to the prediction.
Feature attribution methods are arguably the most studied (and benchmarked) methods in interpretable machine learning, \eg \cite{Ancona2018TowardsNetworks}. Here we take a slightly different perspective from previous work. We posit that the importance of a feature should be proportional to \emph{how imprecise would the prediction be if we did not know its value}. More precisely, let us assume that we are provided with attributions $a_i$ to explain the importances of features $i$ with $i = 1, \dots, N$ for a function assuming value $y^* = f(\mathbf{x}^*)$ at a point $\mathbf{x}^* = (x^*_1, \dots, x^*_N)$. Then, we claim that the following desideratum should hold: 
\begin{equation}\label{eq:attr_prop}
    |a_i| \propto \mathbb{E}(l(y^*, f_{i}) | \mathbf{x}^*_{-i}) = \int_{\mathcal{X}_i} l(y^*, f_i(x_i)) p(x_i)\mathrm{dx_i}
\end{equation}

where $f_i$ is the restriction of the function $f$ to the feature $i$ obtained by fixing the other features at the values $\mathbf{x}^*_{-i} = (x_1,\dots,x_{i-1},x_{i+1}, x_N)$, and $l$ is a performance measure of interest (\eg cross-entropy). With a slight abuse of notation, the probability density in Eq.~\eqref{eq:attr_prop} could either be independent or dependent (\ie conditional) on the values of the other features, depending if the wanted interpretation is \emph{true to the model} or \emph{true to the data}~\cite{Lundberg2017APredictions}. 

%
The desideratum in Eq.~\eqref{eq:attr_prop} naturally leads to two possible metrics.

\begin{metric}[Monotonicity] The \emph{monotonicity} for feature attributions $a_i$ is defined as the Spearman's correlation coefficient $\rho_S(\mathbf{a}, \mathbf{e})$~\cite{Daniel1978AppliedStatistics}. $\mathbf{a} = (\dots,|a_i|,\dots)$ is a vector containing the absolute values of the attributions. $\mathbf{e} = (\dots,\mathbb{E}(l(y^*, f_{i}); X_i | \mathbf{x}^*_{-i}),\dots)$ contains the corresponding (estimated) expectations, as computed in Eq.\eqref{eq:attr_prop}. 
\end{metric}

\begin{metric}[Non-sensitivity] Let $A_0 \subset \{1, \dots, N\}$ be the subset of indeces $i$ denoting features with assigned zero attribution, \ie $a_i = 0$. Further, let $X_0 = \{i \in \{1, \dots, N\}| \mathbb{E}(l(y^*, f_{i}) = 0\}$ denote the subset of indeces $i$ of features to which the model $f$ is not functionally dependent on. The \emph{non-sensitivity} is computed as $|A_0 \triangle X_0|$
where $|\cdot|$ is the cardinality of a set, and $\triangle$ denotes a symmetric difference. 
\end{metric}

Monotonicity and non-sensitivity are indicative of how faithful a feature attribution explanation is. If attributions are not monotonic then we argue that they are not providing the correct importance of the features (Section~\ref{sec:exp_feat_att}). Non-sensitivity ensures that a method assigns zero-importance \emph{only} to features to which the model $f$ is not \emph{functionally} dependent on. This is aligned with the \emph{Sensitivity(b)} axiom proposed by \citet{Sundararajan2017AxiomaticNetworks}. The number of non-zero attributions $|\{1,\dots,N\} \setminus A_0|$ serves as a measure of \emph{complexity}.

Monotonicity, non-sensitivity, and complexity account only for \emph{individual} features. We therefore propose the \emph{effective complexity} to account for feature interactions. 

\begin{metric}[Effective Complexity] Let $a^{(i)}$ be the attributions \emph{ordered increasingly \wrt their absolute value}, and $x^{(i)}$ the corresponding features. Let $M_k = \{x^{(N-k)}, \dots, x^{(N)}\}$ be the set of top $k$ features. Given a chosen tolerance $\epsilon > 0$,
the \emph{effective complexity} is defined as
\begin{equation}\label{eq:min_set}
    k^* = \mathrm{argmin}_{k \in \{1,\dots,N\}} |M_k| \;\; \mathrm{s.t.} \;\; \mathbb{E}(l(y^*, f_{-M_k}) |  \mathbf{x}^*_{M_k}) < \epsilon
\end{equation}
where $f_{-M_k}$, similarly to the definition above, is the restriction of the model $f$ to the non-important features, given fixed values for the (important) features in $M_k$. 
\end{metric}

A low effective complexity means that we can ignore some of the features even though they do have an effect (reduced cognitive salience) because the effect is actually small (non-sensitivity \`a la \citet{Ylikoski2010DissectingPower}). This comes at the cost of reduced factual accuracy, but provides a higher degree of integration since we can freely change the value of the unimportant features without significant effects on the prediction. Therefore explanations with low effective complexity are both simple and broad.

\section{Experimental validation}\label{sec:experiments}

\subsection{Feature extractor effect on LIME}\label{sec:feat_ext_lime}

In this section we analyze the effect of the feature extractor on the local explanation provided by LIME~\cite{Ribeiro2016Classifier}. The target model is a decision tree classifier trained on the \texttt{Avila} dataset~\cite{DeStefano2018ReliableCase}. The dataset consists of 10 normalised features describing the writing patterns of 12 copyists of the \emph{Avila bible} from the \RNum{12} century. The task is to identify the copyist. The tree is trained using the Gini Index~\cite{Gini1912VariabilitaMutabilita} for split selection, and a maximum depth of five to maintain the decision rules easy for comparison. This setting leads to a relatively poor model with test accuracy of 60.38\%. 

\begin{figure}
\centering
\begin{subfigure}{.27\textwidth}
  \centering
  \includegraphics[width=\linewidth]{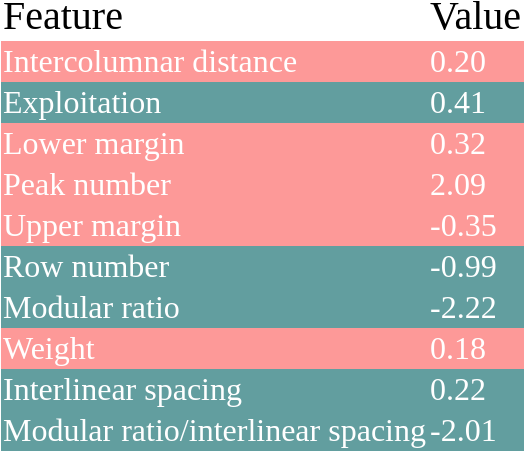}
  \caption{\texttt{Identity}.}
  \label{fig:feat_ext_id}
\end{subfigure}
\hfill
\begin{subfigure}{.27\textwidth}
  \centering
  \includegraphics[width=\linewidth]{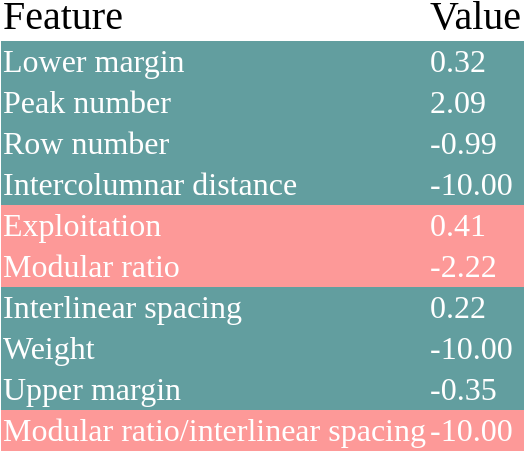}
  \caption{\texttt{Random}.}
  \label{fig:feat_ext_random}
\end{subfigure}
\hfill
\begin{subfigure}{.34\textwidth}
  \centering
  \includegraphics[width=\linewidth]{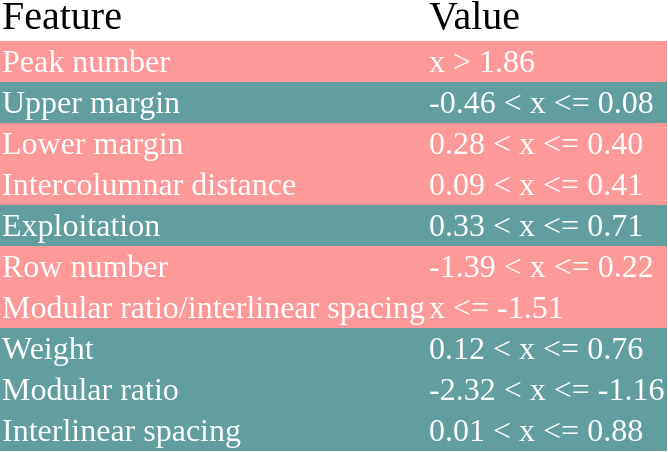}
  \caption{\texttt{Entropy} discretizer.}
  \label{fig:feat_ext_entropy}
\end{subfigure}
\caption{Comparison of feature ranking provided by LIME using 3 different feature extractors. In each Figure, the feature identifier is reported on the left, while its value is reported on the right. The features are reported in order of importance towards the prediction, with top features being the most important. Features highlighted in red (resp. green) constitute positive (resp. negative) evidence.}
\label{fig:extractor_comparison}
\end{figure}

We analyze the local explanation provided for a sample from the test set, which was (mis-)classified as copyist ``7''. The true classification rules of the target tree classifier,  to predict class ``7'' are:
\begin{itemize}
    \item \texttt{Intercolumnar distance} $> -0.10$  AND \texttt{Upper margin} $> -0.03$  AND \texttt{Lower margin} $> 0.17$ AND \texttt{Exploitation} $\leq 0.05$  AND \texttt{Row number} $\leq 0.84$; OR
    \item \texttt{Intercolumnar distance} $> -0.10$ AND $0.05 \leq$ \texttt{Exploitation} $\leq 0.73$ AND \texttt{Row number} $\leq 0.84$  AND \texttt{Peak number} $> 1.11$.
\end{itemize}
For feature extraction we consider \texttt{Identity} (\ie maintaining the original features), \texttt{Random} (\ie assigning the out-of-distribution value $-10.00$ to three randomly selected features), and a discretizer provided in the original implementation of LIME:\footnote{\url{https://github.com/marcotcr/lime/tree/master/lime}} \texttt{Entropy}. This discretizer first trains relatively shallow trees \emph{for each individual feature} using the information entropy criterion. It then discretizes the features using the split learned by such trees. The right-most part of Figure~\ref{fig:feat_ext_entropy} shows an example of this discretization.

In Table~\ref{tab:mi_vs_score} we report the mutual information estimates (using~\cite{Kraskov2004EstimatingInformation}) computed for each feature extractor. The target random variable $Y$ used for the computation of $MI(Z, Y)$ is the prediction generated by the decision tree that we are (locally) interpreting. We also report the $R^2$ score of LIME in approximating the tree prediction probabilities (``7'' vs. ``not-7'') for the sample at hand.   

The \texttt{Identity} extractor expectedly is the best in terms of mutual information between the original features and ``extracted features'': for this case the mutual information just corresponds to the upperbound, \ie the entropy of the original features. Figure~\ref{fig:feat_ext_id} shows the feature rankings generated by LIME using these features. LIME correctly assigns top ranks to all the features used by the target decision tree to classify a sample as copyist ``7''. Further, note how LIME correctly assigns the positivity/negativity of the evidence (color-coded in the figure) in accordance to the true rules, \eg \texttt{Intercolumnar distance}$=0.20$ constitutes positive evidence, in agreement with the rule \texttt{Intercolumnar distance}$>-0.10$.

The \texttt{Random} extractor understandably performs worse in terms of mutual information $MI(X,Z)$ between features. However, the mutual information with the target variable $MI(Z,Y)$ is comparable to the computed for the \texttt{Identity} extractor. This suggests that, \emph{on average}, we lost information contained in the original features, but managed to maintain enough information to potentially replicate the predictions of the target decision tree. The sample we are studying here is, however, an exception. Figure~\ref{fig:feat_ext_random} shows that one of the randomly modified features is the \texttt{Intercolumnar distance}, which is important for the classification of a sample as copyist ``7'', as discussed above. This seems to cause LIME to provide incorrect results, especially in terms of positivity/negativity of the evidence. Interestingly, the approximation $R^2$ score reported by LIME misleadingly suggests that by leveraging a \texttt{Random} extractor, a good (\wrt to the \texttt{Identity} case) explanation can be generated. However, as we just discussed this is not true, which shows how \emph{local approximation scores should not be completely relied on in evaluating a local explanation}. 

Note however that, while a lower value of the mutual information metrics does correspond to an inaccurate explanation, the value itself \emph{may not monotonically correlate with the degree of inaccuracy} of the explanation. This is shown in the case of the \texttt{Entropy} extractor. The discretization introduced by this extractor (shown in Figure~\ref{fig:feat_ext_entropy}) lead to even lower mutual information scores. Nonetheless, LIME managed to provide rules more in agreement with the true rules compared to the \texttt{Random} extractor case. Furthermore, simplifying the original feature through discretization provides a further advantage: many of the original samples actually have the same representation in the discretized space. This means that the explanation is not only \emph{simpler}, but more \emph{broad}.

\begin{minipage}[t]{0.5\textwidth}
  \centering
      \captionof{table}{Mutual Information (MI) vs Accuracy Score of LIME for a local explanation using 5 different feature extractors. Mean values over 50 runs. 
      }\label{tab:mi_vs_score}
\begin{tabular}{lrrr}
\toprule
{} &  $MI(X, Z)$ &  $MI(Z, Y)$ &     $R^2$ \\
\midrule
\texttt{Identity}        &                    8.80 &                   0.49 &  0.13 \\
\texttt{Random}      &                    3.70 &                   0.51 &  0.39 \\
\texttt{Entropy}     &                    0.92 &                   0.41 &  0.09 \\
\bottomrule
\end{tabular}
  \end{minipage}
  \hfill
\begin{minipage}[t]{0.4\textwidth}
  \centering
      \captionof{table}{Metrics for example-based methods: non-representativeness (NR) and diversity (D). Comparison of \texttt{K-medoids}, \texttt{MMD}, and \texttt{ProtoDash}.}\label{tab:example_metrics}
\begin{tabular}{lrr}
\toprule
{} &         NR &          D \\
\midrule
\texttt{K-medoids}   &  0.04 &   8.81 \\
\texttt{MMD}       &  0.30 &   9.59 \\
\texttt{ProtoDash} &  0.70 &  11.01 \\
\bottomrule
\end{tabular}
    \end{minipage}
    
\subsection{Example-based explanations on Rotated MNIST}

\begin{figure}[h]
  \begin{subfigure}[b]{0.32\textwidth}
    \centering
    \scalebox{-1}[1]{\includegraphics[width=\textwidth]{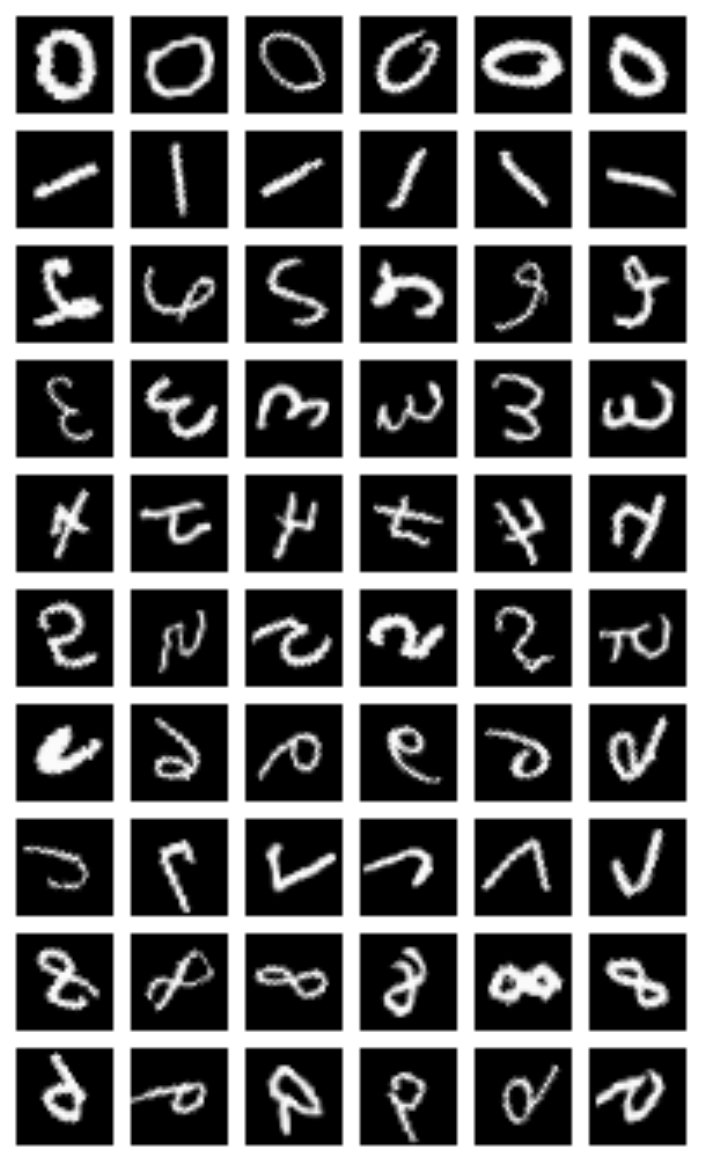}}
    \caption{\texttt{K-Medoids}.}\label{fig:kmedoids}
  \end{subfigure}
  \hfill
  \begin{subfigure}[b]{0.32\textwidth}
    \centering
    \scalebox{-1}[1]{\includegraphics[width=\textwidth]{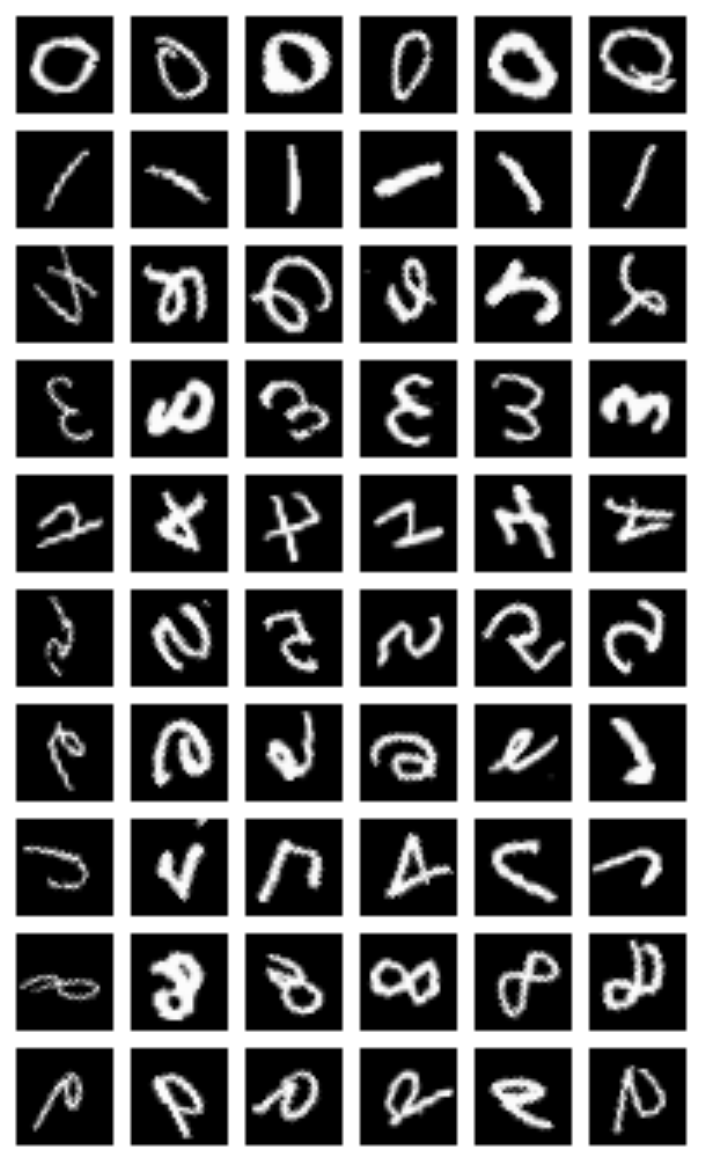}}
    \caption{\texttt{MMD}.}\label{fig:mmd}
  \end{subfigure}
  \hfill
  \begin{subfigure}[b]{0.32\textwidth}
    \centering
    \scalebox{-1}[1]{\includegraphics[width=\textwidth]{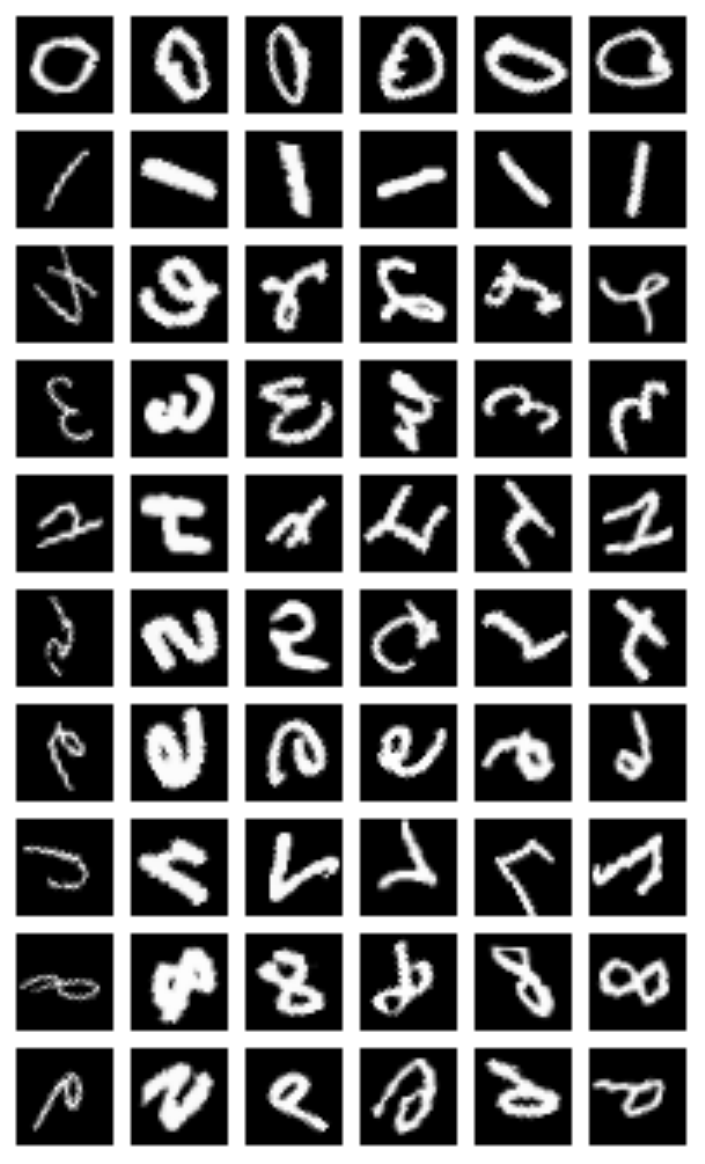}}
    \caption{\texttt{ProtoDash}.}\label{fig:protodash}
  \end{subfigure}
  \caption{Prototypes produced by methods for example-based explanations.}\label{fig:prototypes}
\end{figure}

We showcase the metrics for example-based explanations (Section~\ref{sec:example_based}) on an image classification task. We train a CNN model~\cite{LeCun1998Gradient-basedRecognition} on the Rotated MNIST dataset~\cite{Larochelle2007AnVariation}. The trained model achieves a test accuracy of 92\%. For interpretability, we aim at understanding how the target CNN model performs the classification by producing exemplars/prototypes for each of the ten classes. We consider three different methods: \texttt{K-medoids}~\cite{Kaufman1987ClusteringMedoids}, \texttt{MMD}~\cite{Kim2016ExamplesInterpretability}, and \texttt{ProtoDash}~\cite{Gurumoorthy2017EfficientWeights}. In Figure~\ref{fig:prototypes}, we show the prototypes selected by each method when selecting six prototypes per class. Table~\ref{tab:example_metrics} reports the relative metrics averaged over all classes.

The metrics reveal that \texttt{K-medoids} provides the most representative prototypes, suggesting that the classification rules that can potentially be inferred by looking at these exemplars are the most \emph{faithful} ones. Unfortunately this comes at the cost of lower diversity, as shown by both the diversity metric and Figure~\ref{fig:kmedoids}. This means that the inferred classification rules will not generalize. An example of these claims can be seen in the prototypes for class 4. Looking at the exemplars generated by \texttt{K-medoids}, one may infer that an open upper part and a relatively thin stroke are distinctive features of the digit 4. While these may be high-precision rules, they are not broad since the prototypes produced by the other methods show that digits classified as 4 can have very different strokes and styles, including having a closed upper part (fifth prototype generated by \texttt{MMD}).
Another advantage of more general rules is that they may help reveal the reasons for a relatively low test accuracy. For example, the \texttt{ProtoDash} prototypes suggest that the target CNN may confuse 7 as 2, or 4 as 5.

\begin{figure}
  \begin{subfigure}{0.48\textwidth}
    \centering
    \includegraphics[width=\textwidth, ]{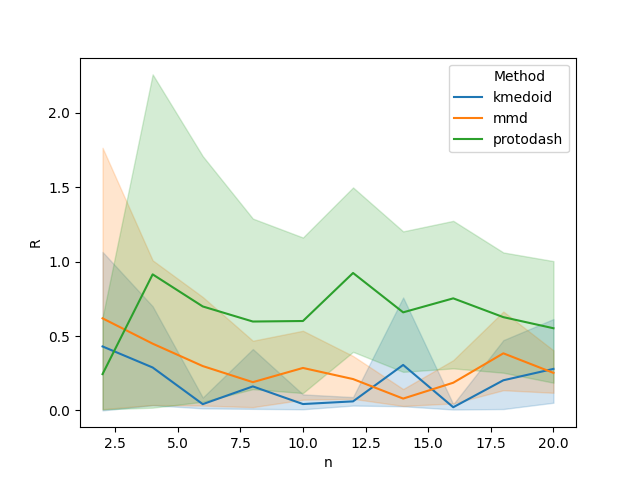}
    \caption{Non-Representativeness (NR).}\label{fig:r_v_n}
  \end{subfigure}
  \hfill
  \begin{subfigure}{0.48\textwidth}
    \centering
    \includegraphics[width=\textwidth]{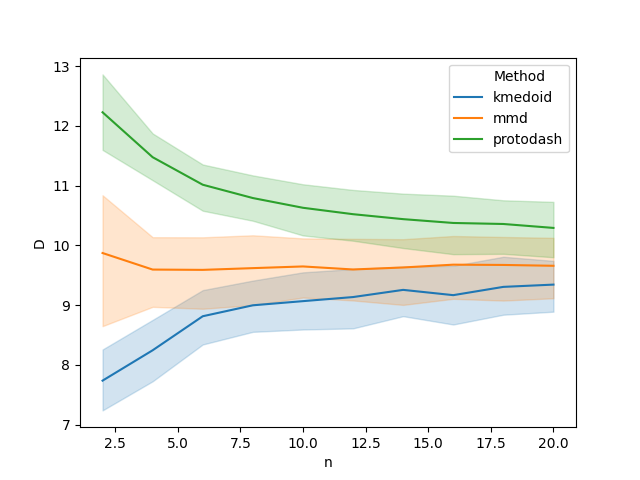}
    \caption{Diversity (D).}\label{fig:d_v_n}
  \end{subfigure}
  \caption{Example-based metrics as functions of the number of exemplars (n).}\label{fig:m_v_n}
\end{figure}

The above discussion has shown how our metrics can be used to assess the fidelity and broadness of example-based explanations. The simplicity of an example-based explanation is simply the number of the exemplars. It is interesting to show how our metrics evolve in relation to the number of examples used (Figure~\ref{fig:m_v_n}). \texttt{ProtoDash} in general seems to be consistently less representative and more diverse. While (expectedly) the diversity for different methods converges the more prototypes are selected, the mean representativeness seem to stay relatively constant. This suggests that, when aiming to have simple explanations, \texttt{ProtoDash} should be preferred for more general explanations, while \texttt{K-Medoids} for more faithful ones.

\subsection{Evaluating feature attributions}\label{sec:exp_feat_att}

We validate our metrics for feature attribution methods on two tasks. First we study how various methods perform in interpreting a non-linear function with \emph{known} analytical formula which will act as a \emph{ground truth}. We then extend our conclusions to a text classification task with higher dimensionality.

\begin{table}[ht]
\centering
\caption{Feature attribution metrics: complexity (C), monotonicity (M), effective complexity (EC), and non-sensitivity (NS). For the Text Classification task we report also the accuracy for a Perturbation Test (PT).}
    \begin{subtable}[t]{.5\textwidth}
    \centering
        \caption{Test Function.}
   \begin{tabular}{lrrrr}
\toprule
              &  C &  M &  EC &  NS \\
\midrule
\texttt{Saliency} &         4 &      0.99 &                     3 &              0 \\
\texttt{InpXGrad} &         4 &      0.82 &                     3 &              0 \\
\texttt{IntGrad} &         4 &      0.59 &                     4 &              0 \\
\texttt{Random} &         6 &      0.12 &                     6 &              2 \\
\bottomrule
\end{tabular}
   \label{tab:test_feat_attr}
    \end{subtable}\hfill
   \begin{subtable}[t]{.5\textwidth}
        \centering
        \caption{Text Classification on News Topics.}
   \begin{tabular}{lrrrrr}
\toprule
         Method &  C&  M & EC & NS &  PT \\
\midrule
\texttt{Saliency} &         25 &      0.00 &                 17 &              0 &                0.77 \\
\texttt{InpXGrad} &         25 &      0.35 &                     13 &              0 &                0.75 \\
\texttt{IntGrad} &         25 &      0.20 &                     6 &              0 &                0.76 \\
\bottomrule
\end{tabular}
   \label{tab:text_class_attr}
    \end{subtable}
\end{table}

\subsubsection{Test Functions}

We showcase our proposed metrics to evaluate feature attribution methods applied to interpret the test function $f: \mathbf{x} \in [0, 1)^6 \mapsto  \frac{2}{3}e^{(x_0 + x_1)} - x_3 \sin (x_2) + x_2$ proposed by \citet{Park1992TuningDesigns} at point $\mathbf{x}^* = (0.24, 0.48, 0.56, 0.99, 0.68, 0.86)$. Note that the function includes two ``non-active'' variables to further test the non-sensitivity of the methods. The methods we consider are: pure gradient, also known as \texttt{Saliency}~\cite{Simonyan2013DeepMaps}, gradients multiplied by the input (\texttt{InpXGrad})~\cite{Kindermans2016InvestigatingNetworks},integrated gradients  (\texttt{IntGrad})~\cite{Sundararajan2017AxiomaticNetworks}. For sanity check we analyze also randomly assigned attributions (\texttt{Random}). The metrics are reported in Table~\ref{tab:test_feat_attr}. The computed attributions are provided in the supplementary for reference (Figure~\ref{fig:attributions}).

As expected, \texttt{Random} performs worse than the other methods on all four metrics: it fails to assign the correct importances leading to a non-monotonic behaviour and to a higher complexity. In particular, it fails to detect the dummy variables. This is not the case for the other considered methods which are non-sensitive to the dummy variables, correctly assigning importance only to the other 4 variables. 

\texttt{IntGrad} performs is less monotonic than other methods. This is because \texttt{IntGrad}, for the sample at hand, assigns less importance to $x_0$ than to $x_2$ and $x_3$, which we argue is incorrect (\ie not faithful). Indeed, for the test function we are analyzing, the contribution of the terms $-x_3 \sin (x_2) + x_2$ is smaller than the exponential term, in the given domain. Therefore, not knowing the value of $x_0$ (or $x_1$) would lead to a higher uncertainty when predicting the value of the test function $f$. For a similar reason, \texttt{InpXGrad} performs worse than \texttt{Saliency}: it assigns less importance to $x_0$ than to $x_3$. 

Finally, let us consider the effective complexity. \texttt{InpXGrad} and \texttt{Saliency} report 3 effectively important variables. We argue that, once again, they provide a more faithful representation of the function $f$. Indeed, for $x_3=0.99$ and $x_2 \in [0, 1)$, we can approximate the non-exponential terms $- x_3 \sin (x_2) + x_2 \approx -0.99x_2 + x_2 = 0.01x_2$, meaning that the contribution of $x_2$ to the function $f$ is effectively negligible in this case. The effective complexity therefore shows that \texttt{InpXGrad} and \texttt{Saliency} provide more broad explanations in the sense discussed in Section~\ref{sec:feat_attr}.

\subsubsection{Text Classification}

We further validate our metrics on a more complex task: text classification. The task consists of classifying news excerpts from the \texttt{AG News Corpus}~\cite{Zhang2015Character-LevelClassification} according to the topic. We train a CNN model reaching 89\% test accuracy. We now want to understand which words from the excerpt are the most important. The sample we picked for this demonstration is shown in the supplementary material (Figure~\ref{fig:text_perturbations}). The performance of \texttt{Saliency}, \texttt{InpXGrad}, and \texttt{IntGrad} according to our metrics is reported in Table~\ref{tab:text_class_attr}.

Our results show that \texttt{IntGrad} provide the most general explanation (low effective complexity), while not being monotonic. This suggests that, for this sample, \texttt{IntGrad} is able to find the most important \emph{group} of words for the classification, but it is not able to accurately estimate the importance of each \emph{individual} word. 

To demonstrate that the explanation provided by \texttt{IntGrad} is the most general, we perform a perturbation test. For each method, we fix the most important words, while substituting the other words with another word randomly picked from the dataset corpus. We then classify the perturbed samples with the trained CNN. The score (PT) for this test, reported in Table~\ref{tab:text_class_attr}, indicates the ratio of perturbed samples that maintained the same predicted class as the original sample.
Note that the number of words kept fixed for each method depends on the identified effective complexity. Our results show that, despite having a higher number of perturbed words, \texttt{IntGrad} performed similarly to the other methods, confirming the broadness of its explanation.

\section{Conclusion and Related Work}

In this work we provide a novel set of metrics for the evaluation of interpretability methods on the basis of objective and quantifiable aspects of interpretability. Key steps in our work are: 
\begin{enumerate*}
    \item the identification of the quantifiable aspects;
    \item the identification of two conceptual parts of an interpretability method, which is necessary for a proper evaluation;
    \item the definition of interpretability metrics.
\end{enumerate*} 
We experimentally validated the metrics on different benchmark datasets and showcased how they can be leveraged to identify the most appropriate for the task at hand. The most appropriate method will depend on the task and on the target user of the explanation: these two aspects will define the trade-off between simplicity, broadness, and fidelity.

To the best of our knowledge, we are the first to propose a systematic accounting of functionally-grounded~\cite{Doshi-Velez2017TowardsLearning} metrics for interpretability evaluation. In particular, we are not aware of any work towards programmatically evaluating the interpretability of the feature extractor and example-based methods. We however acknowledge that previous work did recognize the pivotal role of the feature extractor in the design of an interpretability method~\cite{Doshi-Velez2017TowardsLearning,Ribeiro2016Classifier,Ribeiro2018Explanations,Sokol2019BLIMEy:LIME}. On the other hand, this conceptual separation was not leveraged for evaluation purposes.

Feature attributions are perhaps the most studied methods from a benchmarking perspective. \citet{Ancona2018TowardsNetworks} propose \emph{sensitivity-$n$}. This metric allowed the authors to reach conclusions about the \emph{algorithmic behaviour} of the considered methods. In particular, if \emph{all} the models satisfy sensitivity-$n$ for all $n$, they can conclude that the model they are interpreting behaves linearly. It remains however unclear how to use this metric to compare the methods in terms of interpretability. \citet{Hooker2019ANetworks} propose a re-training strategy, ROAR, as a metric for interpretability evaluation. We argue that a re-trained model would give little insight about the decision process of the original model that was to be interpreted, \ie ROAR is not measuring the fidelity of the attributions. Indeed, if a model re-trained on a dataset with removed feature performs similarly to the original model, it may just mean that the feature the model was using has high correlation with another feature in the dataset: ROAR measures \emph{data} interpretability, rather than \emph{model} interpretability. \citet{Yeh2019OnExplanations} recently proposed an infidelity metric. Interestingly it can be shown that various attribution methods optimize this measure, depending on the distribution used to compute the metric. However, their proposed metric requires the definition of baselines, which we argue are not necessary in general. The metric that perhaps is closest to our proposals is the monotonicity metric proposed by \citet{Arya2019OneTechniques}. However, their monotonicity measure is \wrt a probability drop, while ours is \wrt the uncertainty in probability estimation. Furthermore, depending on the chosen loss function $l$, our definition can be extended to regression tasks. 





\section*{Broader Impact}

The need of interpretability towards societal issues is undeniable, starting from ethical~\cite{Davis2015EthicalSuperintelligence} and safety~\cite{Amodei2016ConcreteSafety} concerns. As discussed in the introduction, despite the rapid growth of the interpretable machine learning field, it is not clear how to compare methods without resorting to user-studies. We argue that this lack of metrics will lead to an \emph{inefficient} development of the field, ultimately working against its rapid growth. Our work is therefore a step forward towards clarifying which objectives should be pursued to ``optimize interpretability'', with consequent advantages to the downstream societal issues. Currently, we do not see any direct negative impact of our work presented here.

\bibliographystyle{unsrtnat}
\bibliography{references}

\clearpage

\renewcommand{\appendixtocname}{Supplementary material}
\section*{Supplementary Material}
\begin{appendices}
  \section{Additional Figures}\label{supp:add_figures}
  
  \begin{figure}[ht]
    \centering
    \includegraphics[width=\textwidth]{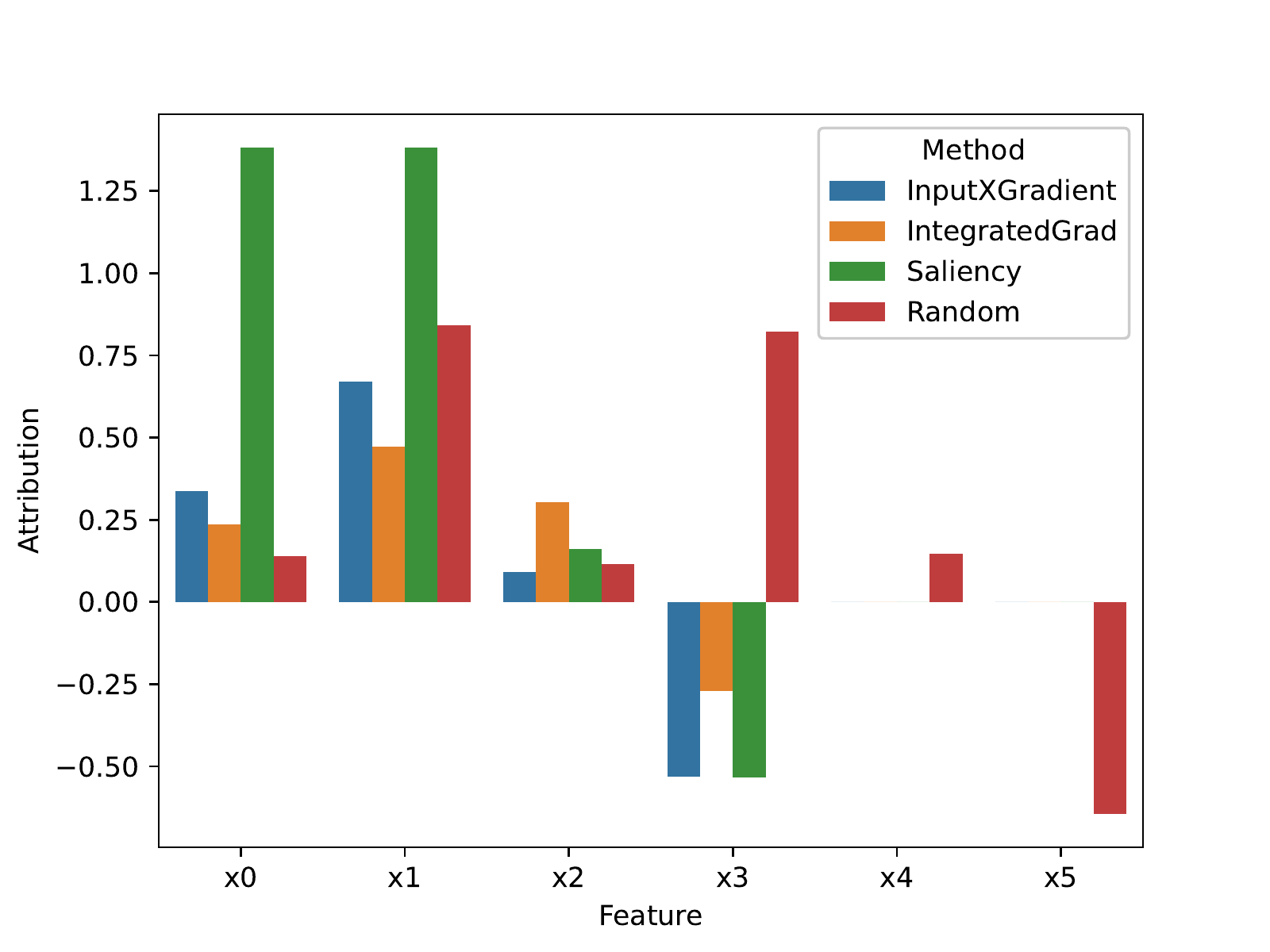}
    \caption{Attributions computed for the Park test function.}\label{fig:attributions}
  \end{figure}
  
\begin{sidewaysfigure}[ht]
  \begin{subfigure}{\textwidth}
    \centering
    \includegraphics[width=\textwidth, ]{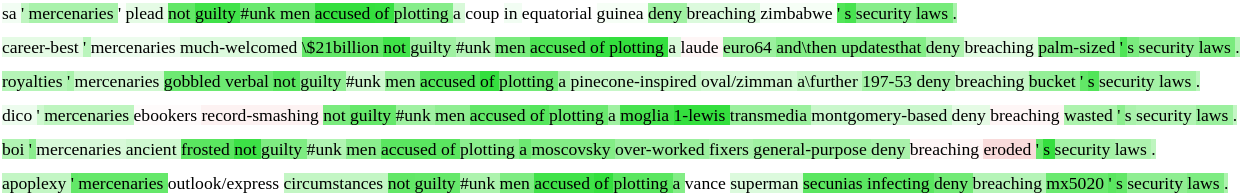}
    \caption{Example of perturbations with attributions provided by \texttt{Saliency}.}
  \end{subfigure}
  
  \begin{subfigure}{\textwidth}
    \centering
    \includegraphics[width=\textwidth]{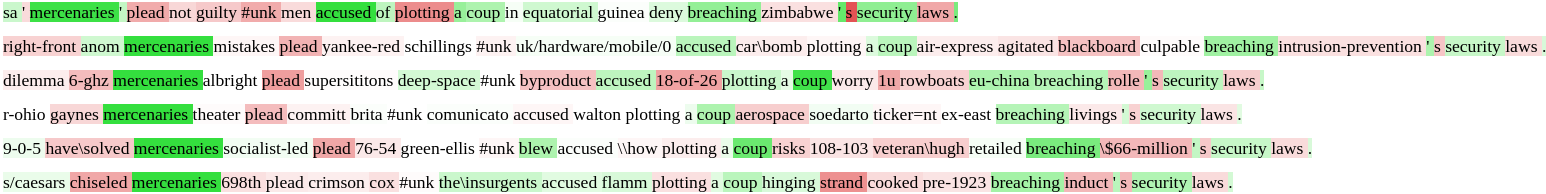}
    \caption{Example of perturbations with attributions provided by \texttt{InpXGrad}.}
  \end{subfigure}
  
  \begin{subfigure}{\textwidth}
    \centering
    \includegraphics[width=\textwidth]{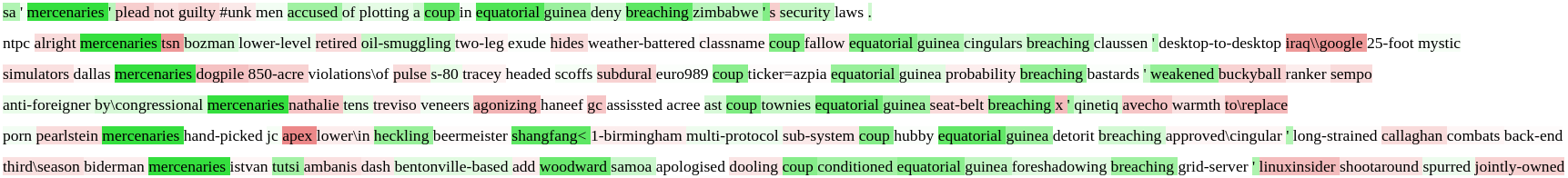}
    \caption{Example of perturbations with attributions provided by \texttt{IntGrad}.}
  \end{subfigure}
  
  \caption{Perturbation samples for each analysed method. The original sample is at the top of each subfigure. Note how the samples for \texttt{IntGrad} contain more perturbed words.}\label{fig:text_perturbations}
  
\end{sidewaysfigure}
  
\end{appendices}

\end{document}